\documentclass{article}

\usepackage{microtype}
\usepackage{graphicx}
\usepackage{subfigure}
\usepackage{booktabs} %

\usepackage{hyperref}

\usepackage[table]{xcolor}

\usepackage[accepted]{icml2023}

\usepackage{amsmath}
\usepackage{amssymb}
\usepackage{mathtools}
\usepackage{amsthm}

\usepackage[capitalize,noabbrev]{cleveref}

\theoremstyle{plain}

\theoremstyle{definition}

\theoremstyle{remark}

\usepackage[textsize=tiny]{todonotes}

\usepackage{siunitx}
\usepackage{multirow}
\usepackage{enumitem}
\usepackage{mathtools}

\newcommand{\fun}[3]{\ensuremath{#1\colon #2\mapsto #3}}

\newcommand{\R}{\mathbb{R}}
\newcommand{\C}{\mathbb{C}}

\newcommand{\SO}{\ensuremath{\mathbf{SO}}}

\newcommand{\timesnarrow}{{\mkern-2mu\times\mkern-2mu}}

\newcommand{\li}[2]{{_#1}{#2}}

\robustify\cellcolor
\newcommand{\bgl}{\cellcolor[HTML]{DDDDDD}}
\newcommand{\bgd}{\cellcolor[HTML]{BBBBBB}}

\defcitealias{klicpera20_fast_uncer_aware_direc_messag}{DimeNet++~\citeyearpar{klicpera20_fast_uncer_aware_direc_messag}}
\defcitealias{tholke2021equivariant}{TorchMD-Net~\citeyearpar{tholke2021equivariant}}
\defcitealias{pmlr-v139-schutt21a}{PaiNN~\citeyearpar{pmlr-v139-schutt21a}} 
\defcitealias{egnn}{EGNN~\citeyearpar{egnn}}
\defcitealias{liao22_equif}{Equiformer~\citeyearpar{liao22_equif}}
\defcitealias{BrandstetterHPB22}{SEGNN~\citeyearpar{BrandstetterHPB22}}

\icmltitlerunning{Scaling Spherical CNNs}

\begin{document}

\twocolumn[
\icmltitle{Scaling Spherical CNNs}

\begin{icmlauthorlist}
\icmlauthor{Carlos Esteves}{google}
\icmlauthor{Jean-Jacques Slotine}{mit}
\icmlauthor{Ameesh Makadia}{google}

\end{icmlauthorlist}

\icmlaffiliation{mit}{Nonlinear Systems Laboratoty, MIT, Cambridge, MA, USA}
\icmlaffiliation{google}{Google Research, New York, NY, USA}
\icmlcorrespondingauthor{Carlos Esteves}{machc@google.com}

\icmlkeywords{Machine Learning, ICML}

\vskip 0.3in
]

\printAffiliationsAndNotice{}  %

\begin{abstract}
Spherical CNNs generalize CNNs to functions on the sphere, by using spherical convolutions as the main linear operation.  The most accurate and efficient way to compute spherical convolutions is in the spectral domain (via the convolution theorem), which is still costlier than the usual planar convolutions. For this reason, applications of spherical CNNs have so far been limited to small problems that can be approached with low model capacity. In this work, we show how spherical CNNs can be scaled for much larger problems. To achieve this, we make critical improvements including novel variants of common model components, an implementation of core operations to exploit hardware accelerator characteristics, and 
application-specific input representations that exploit the properties of our model. Experiments show our larger spherical CNNs reach state-of-the-art on several targets of the QM9 molecular benchmark, which was previously dominated by equivariant graph neural networks, and achieve competitive performance on multiple weather forecasting tasks.
Our code is available \url{https://github.com/google-research/spherical-cnn}.

\end{abstract}
\section{Introduction}\label{sec:intro}
Spherical convolutional neural networks~\cite{s.2018spherical} were introduced as a response to the convolutional neural networks (CNNs) that were central to a series of breakthroughs in computer vision~\cite{krizhevsky2012alexnet,he2016resnet,simonyan2015vgg19,RonnebergerFB15}. Given the prevalence of spherical data across many applications, it seemed sensible to design neural networks that possess attributes analogous to those that contribute to the success of planar CNNs, such as translation equivariance, spatial weight sharing, and localized filters.

Much of the ensuing research into designing spherical CNNs~\cite{s.2018spherical,kondor2018clebsch,EstevesMD20} fulfilled these objectives, providing theoretical guarantees on rotation equivariance, the ability to learn local and expressive filters, and faithful models of both scalar and vector fields on the sphere.

\begin{figure}[t]
  \begin{center}
    \centerline{\includegraphics[width=\columnwidth]{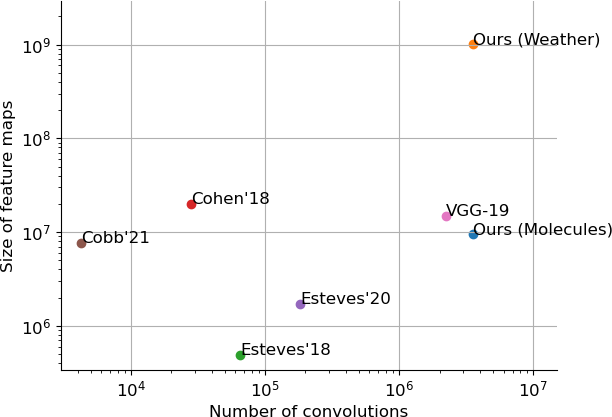}}
    \caption{Previous spherical CNNs were limited to low resolutions
      and relatively shallow models.  In this work, we scale spherical
      CNNs by one order of magnitude and show that they can be
      competitive and even outperform state-of-the-art graph neural
      networks and transformers on scientific applications. In the
      figure, the number of convolutions in a layer mapping between
      $C_{in}$ and $C_{out}$ channels is counted as $C_{in}C_{out}$,
      and the feature map size for a $H \timesnarrow W$ feature with $C$ channels
      is the number of entries $HWC$.
      \vspace{-20pt}
    }
    \label{fig:sizes}
  \end{center}
\end{figure}

Nonetheless, these models have not impacted many real-world applications. One reason is that learning from large datasets requires models with adequate representational capacity, and it has not yet been shown that spherical convolution layers can be composed to construct such large models effectively.  See Figure~\ref{fig:sizes} -- there is no spherical CNN architecture used in practice analogous to common CNN models such as VGG19~\cite{simonyan2015vgg19}.

We are inspired by scientific applications in two areas, drug discovery and climate analysis, that have the potential for broad societal impact. Naturally, both have drawn great interest from the machine learning community -- for example AlphaFold for predicting 3D structure of proteins~\cite{AlphaFold2021}, and a litany of deep learning approaches for molecular property prediction~\cite{wieder2020molecular}.
Property prediction of small molecules may also be relevant in the design of drugs targeting the interaction between two proteins.
For instance, current cancer drugs based on disrupting the binding of tumor suppressor p53 and ubiquitin ligase MDM2 (which targets p53 for degradation) have very low efficiency~\cite{SUN2006645}.
The second area of interest is short and medium-range weather forecasting~\cite{ravuri2021skilful,graphcast,rasp20_weath}.
Climate interventions are being considered  to mitigate the effects of increased greenhouse gas concentrations in the atmosphere\footnote{\url{https://www.ametsoc.org/index.cfm/ams/about-ams/ams-statements/statements-of-the-ams-in-force/climate-intervention}}.
As such
interventions represent uncharted and potentially dangerous territory, climate prediction models may prove important to improve
their safety and effectiveness.

Intuitively, both molecular property prediction and climate forecasting problems should benefit from spherical CNNs.  The intrinsic properties of molecules are invariant to rotations of the 3D structure (atom positions), so representations that are rotation equivariant by design would provide a natural way to encode this symmetry.  However, QM9~\cite{ramakrishnan2014quantum}, a current standard benchmark for this problem, contains 134K molecules, over 18 times larger than the dataset existing spherical CNNs can accommodate~\cite{rupp2012fast}. The scale of this problem necessitates models with much greater representation power and computational efficiency.

Similarly, climate forecasting datasets~\cite{rasp20_weath} represent samples of the Earth's atmospheric state and thus are ideally represented as spherical signals.
Furthermore, in meaningful forecasting applications, models will rely on numerous input variables and the objective is to predict at a high spatial resolution (e.g.\ \ang{1} angular resolution or 64k samples). Such input and output sizes demand large models.

In this work we present a systematic and principled approach to scale spherical CNNs. Our contributions include
\begin{itemize}[noitemsep,topsep=-0.5\parskip]
\item a design of large scale spherical CNN models, which includes an efficient implementation of spin-weighted spherical harmonic transforms tailored to TPUs,
\item general purpose layers and activations that improve expressivity and efficiency,
\item application-specific modeling for molecules and weather forecasting.
\end{itemize}

As the contributions listed above hint at, we observe a naive scaling of existing spherical CNN architectures (simply increasing depth and/or width) is insufficient. Rather, our larger models required a measured design that altered multiple standard components such as the nonlinearity, batch normalization, and residual blocks -- all of these improved both efficiency and test performance (see \cref{tab:ablation}).

These advancements, along with novel domain-specific input feature representations, lead to state of the art performance on the QM9 benchmark, which has been mostly dominated by variations of graph neural networks and transformers. %
Our models are also competitive in multiple weather forecasting settings,
showing, for the first time, that spherical CNNs are viable neural weather models.

This work shows the feasibility of, and introduces best practices for, scaling spherical CNNs.  Based on our findings, we expect our JAX~\cite{jax2018github} implementation
will provide a platform for further research with spherical CNNs targeting real world applications.

\section{Related work}
\subsection{Spherical CNNs}
Spherical CNNs have been introduced as the natural extension of
standard CNNs to the sphere~\cite{s.2018spherical,esteves18eccv}, with
spherical convolutions computed via generalized Fourier
transforms, where the translation equivariance is generalized to 3D
rotation equivariance.
Later work introduced spectral
nonlinearities~\cite{kondor2018clebsch},
and extended the equivariance to conformal transformations~\cite{MitchelAKK22}.

One set of approaches applies filters directly to discrete samples
of the spherical input. \citet{perraudin2019deepsphere} used a
rotation equivariant graph CNN based on isotropic filters.
\citet{CohenWKW19} considered charts of an icosahedral grid,
where filters are rotated and shared, yielding approximate rotation
equivariance.
\citet{ShakerinavaR21} generalized \citet{CohenWKW19} to other grids,
using less constrained filters that maintain equivariance.

Tangentially related are methods that operate on the sphere but are not
rotation-equivariant \cite{Coors_2018_ECCV,JiangHKPMN19,SuG19}.

There have been attempts at improving spherical CNN's efficiency.
\citet{EstevesMD20} introduced spin-weighted spherical CNNs,
which brought anisotropic filters at smaller cost than the full rotation group
Fourier transforms. %
\citet{cobb2021efficient} counteracted the feature size expansion
caused by the tensor products in~\citet{kondor2018clebsch} with heuristics
to select the representation type at each layer.
\citet{mcewen2022scattering} handled high resolution spherical signals
by first applying a scattering network with fixed filters (not trainable),
followed by downsampling and a spherical CNN.
\citet{ocampo22_scalab_equiv_spher_cnns_by} approximated the group convolution integral
using a quadrature rule, avoiding expensive generalized Fourier transforms.
These previous attempts were still limited to small and
sometimes contrived applications.

In this paper, we scale spherical CNNs
to a number of operations and feature resolutions one order of magnitude
larger than prior work (see \cref{fig:sizes}), and apply them
successfully to large benchmarks on molecule and
weather modeling, showing that they can be comparable and sometimes
surpass the state-of-the-art graph and transformer-based models.
\subsection{Deep learning for molecules}
There have been many flavors of message-passing graph neural networks designed specifically for molecules. See \citet{wieder2020molecular} and \citet{han2022geometrically} for relevant surveys, and a broader discussion not limited to deep learning techniques can be found in~\citet{vonLilienfeld2020}.  Regarding the deep learning approaches, much of the recent work has focused on 3D equivariant or invariant models.

Examples of invariant models include SchNet~\cite{schutt2017schnet} and DimeNet++~\cite{klicpera20_fast_uncer_aware_direc_messag}, where the update function only uses invariant information such as bond angles or atomic distances. E(n)-equivariant networks~\cite{egnn} propose an equivariant update function for node coordinates that operates on directional information (displacement between atoms), although the model instantiation for molecules skips this update leading to strict invariance. Related, Tensor Field Networks~\cite{tensorfieldnets} also construct an equivariant update function, this one is based on spherical harmonics. Cormorant~\cite{anderson2019cormorant} and Steerable E(3)-equivariant GNNs~\cite{BrandstetterHPB22} can be seen as extensions of TFN, the former noted for using the Clebsch-Gordan non-linearity, and the latter generalizing to E(n) equivariance. PaiNN~\cite{pmlr-v139-schutt21a} is a related model whose gated equivariant update block does not rely on spherical harmonics. This work also closely examines the loss of directional information in invariant models and finds equivariance allows for models with reduced model size. Related to the equivariant graph models are the equivariant transformer approaches such as TorchMD-Net~\cite{tholke2021equivariant} which updates scalar and vector node features with self-attention.

\subsection{Deep learning for weather modeling}

\citet{mudigonda2017segmenting} and \citet{weyn2019cnn} utilize vanilla CNNs for extreme weather segmentation and short-term forecasting (``NowCasting''), respectively, while \citet{rasp21_data_medium_weath_predic_with} uses a ResNet model. 
Other methods for NowCasting use UNets~\cite{agrawal2019machine} and conditional generative modeling~\cite{ravuri2021skilful}.

In \citet{weyn2020cubemap} and \citet{GlobalExtremeHeatForecastingUsingNeuralWeatherModels}, a cubed sphere representation (projecting the sphere onto six planar segments) is proposed. This approach enjoys some of the computational benefits of traditional CNNs while more closely observing the underlying spherical topology.
\citet{JiangHKPMN19} introduces a model for unstructured grids with experiments on extreme weather segmentation on icosahedral grids. This orientable CNN model does not offer equivariance to 3D rotations and thus expects inputs to be consistently oriented which is true of climate data. 

\citet{keisler22_forec_global_weath_with_graph_neural_networ} recently introduced a graph neural network model inspired by~\citet{battaglia18relational}. The central component of this model is a message passing network operating on an icosahedral grid. A similar approach is taken in \citet{graphcast}, with a multi-scale mesh graph representation.

The datasets used in most of the recent deep learning research for climate modeling, such as WeatherBench~\cite{rasp20_weath}, consists of equiangular grids derived from reanalysis of the ERA5 data~\cite{era5}.  

\section{Background}
\textbf{Spherical CNNs.}
The ubiquitous convolutional neural networks (CNNs) for image analysis
have convolutions on the plane as their main operation,
\begin{align*}
  (f * k)(x) = \int\limits_{t\in \R^2} f(t)k(x-t)\, dt,
\end{align*}
for an input function $f$ and learnable filter $k$. This operation brings
filter sharing between different regions of the image via translation
equivariance, which means that given a shifted input $f'(x) = f(x + h)$,
the convolution output also shifts: $(f' * k)(x) = (f * k)(x + h)$.
This is one of the main reasons for CNNs high performance.

Spherical CNNs generalize this notion to functions on the sphere
(\fun{f}{S^2}{\R}) by using spherical convolutions, 
\begin{align}
(f * k)(x) = \int\limits_{g \in \SO(3)} f(g \nu) k(g^{-1} x) \, dg, \label{eq:sphconv}
\end{align}
where \SO(3) is the group of 3D rotations
(which can be represented by special orthogonal $3\times 3$ matrices),
and $\nu \in S^2$ is a fixed point.
Any two points on the sphere $S^2$ are related by a rotation in 3D
(in technical terms, the sphere is a homogeneous space of the group of rotations), 
and the spherical convolution is equivariant to 3D rotations.
\cref{eq:sphconv} was adopted by \citet{esteves18eccv}, while
\citet{s.2018spherical} lifts from $S^2$ to $\SO(3)$ and performs
convolutions on the group, which have an almost identical expression
to \cref{eq:sphconv} but with $x\in \SO(3)$.

\citet{EstevesMD20} introduced spin-weighted spherical CNNs to
overcome the limited expressivity of \citet{esteves18eccv} and the
computational overhead of \citet{s.2018spherical}. Spin-weighted spherical functions are
complex-valued and their phase changes under rotation. They can also
be interpreted as functions on $\SO(3)$~\cite{Boyle_2016} with sparse spectrum. 
Convolution is computed through products of spin-weighted spherical
harmonics coefficients~\cite{EstevesMD20}.

\textbf{Computing generalized convolutions.}
Approximating spherical and rotation group convolutions with a discrete sum is
problematic because there is no arbitrarily dense self-similar grid
on the sphere and on the rotation group.
Hence, these convolutions are most efficiently
and accurately computed in the spectral domain, via products of (generalized) Fourier
coefficients~\cite{driscoll1994computing,kostelec2008ffts,Huffenberger_2010},
which correspond to the spherical harmonics decomposition
coefficients $\hat{f}^\ell_m$ for degree $\ell$ and order $m$ in the case of
\cref{eq:sphconv}.
Even the fastest algorithms for spherical and rotation group
convolutions are still much slower than a planar convolution
with a $3\times 3$ kernel on modern devices, which has so far limited the
applications of spherical CNNs.

In this paper, we adapt the algorithm of \citet{Huffenberger_2010} for computing
spin-weighted transforms,
\begin{align}
  \li{s}{\hat{f}_m^{\ell}} &= \int\limits_{S^2} f(x) \overline{\li{s}{Y_m^{\ell}}}(x)\, dx \label{eq:swsft}, \\
  f(x) &= \sum_{\ell} \sum_{|m| \le \ell}\li{s}{\hat{f}_m^{\ell}}\li{s}{Y_m^{\ell}}(x) \label{eq:iswsft},
\end{align}
where $\li{s}{Y_m^{\ell}}$ is the spin-weighted spherical harmonic of spin $s$,
degree $\ell$, and order $m$, and $\li{0}{Y_m^{\ell}}$ corresponds to the
standard spherical harmonic $Y_m^{\ell}$.
The forward transform implementation rewrites \cref{eq:swsft} as 
\begin{align}
  \li{s}{\hat{f}}_m^\ell = (-1)^si^{m+s} \sqrt{\frac{2\ell + 1}{4\pi}}
    \sum_{m'=-\ell}^\ell \Delta_{m'm}^\ell \Delta_{m'(-s)}^\ell
  I_{m'm},
  \label{eq:swsft_rewr}
\end{align}
where $\Delta^\ell_{m,m'} = d^\ell_{m,m'}(\pi/2)$ is 
the Wigner $\Delta$ function,
$d^\ell$ is a Wigner (small) $d$ matrix, and $I_{m'm}$ is an inner product
with $f$ over the sphere, computed by extending $f$ to the
torus and evaluating standard fast Fourier transforms (FFTs) on it.
Symmetries of the Wigner $\Delta$
enable rewriting the sum in \cref{eq:swsft_rewr} with
half the number of terms as
\begin{align}
  \sum_{m'=-\ell}^\ell \Delta_{m'm}^\ell \Delta_{m'(-s)}^\ell I_{m'm} &= \sum_{m'=0}^\ell \Delta_{m'm}^\ell \Delta_{m'(-s)}^\ell J_{m'm},
\label{eq:jnm}                                               
\end{align}
where $J_{m'm} = I_{m'm} + (-1)^{m+s}I_{-m'm}$ for $m' > 0$ and $J_{0m}=I_{0m}$.

The inverse transform rewrites \cref{eq:iswsft} as
\begin{align}
f(\theta, \phi) &= \sum_{m'=-\ell}^\ell \sum_{m=-\ell}^\ell e^{im'\theta}e^{im\phi}G_{m'm}, \\
  G_{m'm} &= (-1)^si^{m+s}\sum_{\ell} \alpha_\ell \Delta^\ell_{(-m')(-s)}\Delta^\ell_{(-m')m} \li{s}{\hat{f}^\ell_m},
\label{eq:gnm} 
\end{align}
where $\alpha_\ell = \sqrt{\frac{2\ell+1}{4\pi}}$.
Again, the Wigner $\Delta$ symmetries imply $G_{m'm} =
(-1)^{m+s}G_{(-m')m}$ so the full $G$ is reconstructed by
computing only half of its values.

The algorithm just described was adopted and implemented in TensorFlow~\cite{tensorflow}
with no changes by \citet{EstevesMD20}. In this work, we offer a
complete rewrite in JAX, tuned for TPUs.  This is by itself faster, but we
also propose modifications to the algorithm to further improve its
speed (see \cref{sec:tpu}).

\section{Method}
We contribute a fast implementation of spin-weighted spherical CNNs in
JAX, optimized for TPUs, that can run distributed in dozens of
devices. The implementation is about $3\timesnarrow$ faster than the original,
and the ability to run distributed can speed it up $100\timesnarrow$ or more
(we use up to 32 TPUs).

Moreover, we introduce a new nonlinearity, normalization layer, and
residual block architecture that are more accurate and efficient than
the alternatives. \cref{tab:ablation} summarizes the effects on
efficiency and accuracy.

\begin{table}[t]
  \caption{
    Effects of our modeling and implementation contributions.
Differences are shown with respect to the results of the previous row.
A model similar to the one described in \cref{sec:qm9} for
enthalpy of atomization on QM9 was used for this analysis.
  }
  \label{tab:ablation}
  \begin{center}
    \begin{tabular}{@{}l
      S
      S
      @{}}
      \toprule
                          & {$\Delta$ Steps/s [\%] $\uparrow$} & {$\Delta$ RMSE [\%] $\downarrow$} \\
      \midrule
      JAX implementation  & +33.7                  & 0.0                 \\      
      Phase collapse      & -4.6                   & -8.0                \\
      No $\Delta$ symmetries   & +16.3                  & 0.0                 \\
      Use DFT             & +21.4                  & 0.0                 \\
      Spectral batch norm & +7.8                   & -1.4                \\
      Efficient residual  & +19.3                  & -2.4                \\
      \bottomrule
    \end{tabular}
  \end{center}
\end{table}

\subsection{Modeling}
\textbf{Phase collapse nonlinearity.}
Designing equivariant nonlinearities for equivariant neural networks containing
vector or tensor features is challenging. A number of equivariant activations
appear in the literature
\cite{weiler20183d,kondor2018clebsch,HaanWCW21,XuLDD22} and typically
the best performing one is problem-dependent.
Spin-weighted spherical CNNs require specialized activations for
nonzero spin features, and \citet{EstevesMD20} chose a simple
magnitude thresholding.

\citet{guth21_phase_collap_neural_networ} introduced phase collapse nonlinearities
for complex-valued planar CNNs with wavelet filters, motivated by
1) translation invariance is usually desirable for image classification,
2) in the spectral domain, translations correspond to phase shifts,
3) when applying complex wavelet filters to images, which yields complex feature maps,
input translations approximately correspond to feature phase shifts,
4) using the modulus as part of the activation collapses the phase, achieving translation invariance and increasing class separability.

We adapt these ideas to spin-weighted spherical functions;
in our case, we want rotation invariance (or equivariance) for functions on the sphere.
The features are complex-valued and an input rotation results
in phase shifts when the spin number is nonzero.
Thus, using the modulus as part of the activation eliminates these shifts and
brings some degree of invariance.
The activation mixes all spins but only updates the zero-spin features.
Since the nonzero spin features are unaffected, no information is lost by collapsing the phase.

Formally, let $x_0 \in \C^C$ be the stack of $C$ channels
of zero spin at some position on the sphere, and $x \in \C^{SC}$ be a
stack of all $S$ spins in the feature map (including zero). We apply
\begin{align*}
  x_0 \leftarrow W_1x_0 + W_2|x| + b,
\end{align*}
where $W_1$, $W_2$ and $b$ are learnable parameters. This operation
updates only the spin zero; subsequent convolutions propagate
information to the nonzero ones. %
\cref{tab:ablation_molecules} shows this nonlinearity brings sizeable
performance improvements.

\textbf{Spectral batch normalization.}
Previous spherical CNNs computed spherical convolutions on the
spectral domain and batch normalization on the spatial domain. Batch
normalization requires approximating the statistics with a quadrature
rule in the spatial domain.  Moreover, in the spin-weighted case, zero
and nonzero spins need different treatments, which is inefficient.

We propose to compute the batch normalization in the spectral domain
instead. Consider that 1) the coefficient $\li{0}{f}^\ell$ for $\ell=0$
corresponds to the function average, and 2) the variance of the rest
of the coefficients is the variance of the function. The normalization
is then computed by 1) setting $\li{0}{f}^0$ to zero and 2) dividing
all coefficients by the variance. Similarly, a learnable bias is
applied by directly setting $\li{0}{f}^0$, and a learnable scale is
applied to all coefficients.

The spectral batch norm is shown to be faster and more accurate than the spatial one in \cref{tab:ablation}. It also enables a faster residual block as described next.
 
\textbf{Spectral pooling and efficient residual block.}
In contrast with \citet{EstevesMD20}, and similarly to
\citet{s.2018spherical}, we perform pooling in the spectral domain,
which proves to be faster and more accurate.
This is because the spatial pooling is sensitive to the sampling grid
so it is only approximately rotation-equivariant;
it also requires approximation with quadrature weights which adds to the errors.
Spectral pooling is implemented by simply skipping the computation of
the higher frequency coefficients within a
spin-spherical Fourier transform.
Spectral pooling is also conceptually different than spatial
because it is a global operation while spatial pooling is localized.

\begin{figure}[ht]
  \begin{center}
    \centerline{\includegraphics[width=\columnwidth]{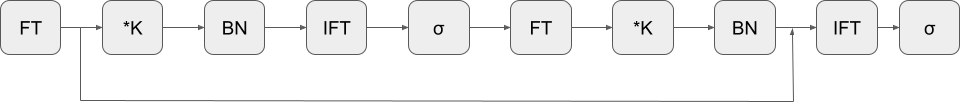}}
    \caption{Our efficient residual block contains spin-weighted spherical
      Fourier transforms (FT) and inverses (IFT), multiplication with
      filter coefficients ($*K$), activation ($\sigma$) and spectral batch
      normalization (BN). The residual connection happens in Fourier
      space. Optionally, spectral pooling is performed at the first FT
      block.
      \vspace{-25pt}
    }
    \label{fig:residual}
  \end{center}
\end{figure}

One potential issue with spectral pooling is in residual layers,
where the downsampling happens in the first Fourier transform,
so the downsampled spatial input is never
computed and hence cannot be used in the skip-connection.  Our
solution is to add the residual connection between Fourier
coefficients, which is enabled by the spectral batch normalization
described earlier. \cref{fig:residual} shows our residual block.
\cref{tab:ablation} shows it is faster and performs better than the
alternative with spatial pooling and batch norms.

\subsection{Efficient TPU implementation}
\label{sec:tpu}
We implement the spin-weighted spherical harmonics transforms aiming
for fast execution on TPUs~\cite{TPUprofiling}\footnote{The implementation is compatible with GPUs as well.}.
This drives our design decisions and
sometimes departs from the optimal implementation for CPUs as
introduced by \citet{Huffenberger_2010}. The main difference is that
TPUs perform matrix multiplications extremely fast, but memory
manipulations like slicing and concatenating tensors may
quickly become a bottleneck.

In particular, the use of Wigner $\Delta$ symmetries to reduce the number
of elements computed in \cref{eq:swsft_rewr,eq:gnm} requires slicing,
modifying and reconstructing the original tensors in order to cut the
number of operations in half. It turns out this is slower on TPU than
just computing twice as many operations without the intermediate steps
for the architecture we consider, so we skip the computation of
$J_{nm}$ (\cref{eq:jnm}) completely, and compute all entries of
$G_{nm}$ (\cref{eq:gnm}) in a single step.

Furthermore, \citet{Huffenberger_2010} leverage the Fast Fourier
transform (FFT) algorithm to reduce asymptotic complexity of the
standard Fourier transforms (from $\mathcal{O}(n^2)$ to $\mathcal{O}(n \log n$)). While
there are on-device implementations of the FFT, it turns out that in
our cases it is significantly faster to compute Fourier transforms as
matrix multiplications via the discrete Fourier transform (DFT)
matrix. This is because, in a typical neural network pass, we will
compute thousands of Fourier transforms (one for each channel for each
convolution), but the resolution of each transform is relatively small
(up to $n=256$ in our experiments), so the constant terms dominate and
there is no benefit in reducing the asymptotic complexity.
\cref{tab:ablation} quantifies the efficiency increase of these changes.

\section{Experiments}

\begin{table*}[t]
  \caption{
    QM9 mean average errors (MAE).
    We scale spherical CNNs for QM9 for the first time,
    and show they are competitive with the previously dominant
    equivariant graph neural networks and transformers.
    We compare on two   splits found in the literature,
    where ``Split 1'' has a larger training set.
    Our model outperforms the baselines on 8 out of 12 targets in ``Split 1''
    and 9 out of 12 targets in ``Split 2''.
    \vspace{-10pt}
}
  \label{tab:qm9}
  \begin{center}
    \resizebox{\textwidth}{!}{
    \begin{tabular}{@{}llc
      S[table-format=1.3,table-auto-round]
      S[table-format=1.3,table-auto-round]
      S[table-format=2.1,table-auto-round]
      S[table-format=2.1,table-auto-round]
      S[table-format=2.1,table-auto-round]
      S[table-format=1.3,table-auto-round]
      S[table-format=1.2,table-auto-round]
      S[table-format=1.2,table-auto-round]
      S[table-format=1.2,table-auto-round]            
      S[table-format=1.2,table-auto-round]
      S[table-format=1.2,table-auto-round]
      S[table-format=1.3,table-auto-round]            
      }
      \toprule
                               &                                       &  & {$\mu$}       & {$\alpha$}       & {$\epsilon_{\text{HOMO}}$} & {$\epsilon_{\text{LUMO}}$} & {$\epsilon_{\text{gap}}$} & {$<R^2>$}   & {zpve}      & {$U_0$}     & {U}          & {H}         & {G}         & {$C_v$}                               \\
                               &                                       &  & {[$D$]}     & {[$a_0^3$]} & {[meV]}             & {[meV]}             & {[meV]}            & {[$a_0^2$]} & {[meV]}     & {[meV]}     & {[meV]}      & {[meV]}     & {[meV]}     & {[$\frac{\text{cal}}{\text{mol K}}$]} \\      
      \midrule
      \multirow{4}{*}{Split 1} & \citetalias{klicpera20_fast_uncer_aware_direc_messag} &  & 0.0297      & \bgd 0.0435 & 24.6                & 19.5                & \bgl 32.6          & 0.331       & \bgl 1.21   & 6.32        & 6.28         & 6.53        & 7.56        & \bgl 0.023                            \\
                               & \citetalias{pmlr-v139-schutt21a}      &  & \bgl 0.012  & \bgl 0.045  & 27.6                & 20.4                & 45.7               & 0.066       & 1.28        & \bgl 5.85   & \bgl 5.83    & \bgl 5.98   & \bgl 7.35   & 0.024                                 \\
                               & \citetalias{tholke2021equivariant}    &  & \bgd 0.011  & 0.059       & \bgd 20.3           & \bgd 17.5           & 36.1               & \bgl 0.033  & 1.84        & 6.15        & 6.38         & 6.16        & 7.62        & 0.026                                 \\
                               & Ours                                  &  & 0.0163      & 0.0489      & \bgl 21.5599        & \bgl 18.0094        & \bgd 28.7863       & \bgd 0.0269 & \bgd 1.1499 & \bgd 5.6539 & \bgd  5.7154 & \bgd 5.6886 & \bgd 6.5353 & \bgd 0.0224                           \\
      \midrule
      \multirow{3}{*}{Split 2} & \citetalias{egnn}                     &  & 0.029       & 0.071       & 29                  & 25                  & 48                 & \bgl 0.106  & 1.55        & 11          & 12           & 12          & 12          & 0.031                                 \\
                               & \citetalias{BrandstetterHPB22}        &  & 0.023       & 0.060       & 24                  & 21                  & 42                 & 0.660       & 1.62        & 15          & 13           & 16          & 15          & 0.031                                 \\
                               & \citetalias{liao22_equif}             &  & \bgd 0.014  & \bgl 0.056  & \bgd 17             & \bgd 16             & \bgl 33            & 0.227       & \bgl 1.32   & \bgl 10     & \bgl 11      & \bgl 10     & \bgl 10     & \bgl 0.025                            \\
                               & Ours                                  &  & \bgl 0.0173 & \bgd 0.0491 & \bgl 22.2765        & \bgl 19.1454        & \bgd 29.7654       & \bgd 0.0282 & \bgd 1.1879 & \bgd 5.9585 & \bgd 5.9822  & \bgd 5.9721 & \bgd 6.9677 & \bgd 0.0233                           \\
      \bottomrule
    \end{tabular}
    }
  \end{center}
  \vspace{-15pt}
\end{table*}

\subsection{Molecular property regression}
\label{sec:molecules_experiments}
We first demonstrate scaling spherical CNNs for
molecular property regression from atoms and their positions in space, a task
that was so far dominated by rotation equivariant graph neural
networks and transformer-based models \cite{wieder2020molecular,klicpera20_fast_uncer_aware_direc_messag,liao22_equif,tholke2021equivariant}. Previous
applications of spherical
CNNs~\cite{s.2018spherical,kondor2018clebsch,cobb2021efficient}
considered only the QM7~\cite{rupp2012fast} dataset, which has \num{7165}
molecules with up to 23 atoms, and a single regression target.
However, the much larger QM9~\cite{ramakrishnan2014quantum}
dataset, which contains \num{134000} molecules with up to 29 atoms and 12 different
regression targets, has supplanted QM7 as the standard benchmark for this task.
The molecules are described by their atom types and 3D positions,
and labeled with geometric, energetic, electronic, and thermodynamic properties
such as enthalpies and free energies of atomization.

In this section, we report the first results of spherical CNNs on
QM9. The main reason to employ spherical CNNs for this task is their
equivariance to 3D rotations, since the molecule properties do not
change under rotations. A secondary reason is that we can design rich
physically-based features when mapping from molecule to sphere.

\subsubsection{Spherical representation of molecules}

The first step for applying spherical CNNs is to
represent the molecule as spherical
functions. \citet{s.2018spherical} proposed a map where spheres
are placed around each atom, and points on each sphere are
assigned a Coulomb-like quantity using the charge of the central atom
and the distances between points on the sphere and other atoms.

We propose an alternative formulation which performs better in practice
(see \cref{tab:ablation_molecules}). Our spherical
functions have no assigned radius, so they only contain directional
information. The values of these functions are constructed from an
inverse power law computed from pairs of atoms, spread out with a
Gaussian decay. The input consists of one set of features per atom, with
one channel per atom type in the dataset. We sum the contributions of
all atoms of the same type. Formally, let $z_i$ be the atomic number
of atom $i$ and $r_{ij}$ the displacement between atoms $i$ and
$j$, we define the one input channel of atom $i$ corresponding to atom
type $z$ as%
\begin{align*}
  f_{iz}(x) = \sum_{z_j = z}\frac{z_iz_j}{|r_{ij}|^p} e^{-\frac{(x \cdot  r_{ij})^2}{\beta|r_{ij}|}}, 
\end{align*}%
where $\beta$ and $p$ are hyperparameters. We set $\beta$ such that the value
is reduced by $95\%$ at \ang{45} away from $r_{ij}$.  We stack the
features for $p=2$ and $p=6$, which correspond to the power laws of
Coulomb and van der Waals forces, respectively. These powers have
been shown to perform well by \citet{huang_understanding_2016} and we
confirm their findings in our setting.

\begin{figure}[t]
  \begin{center}
    \centerline{\includegraphics[width=\columnwidth]{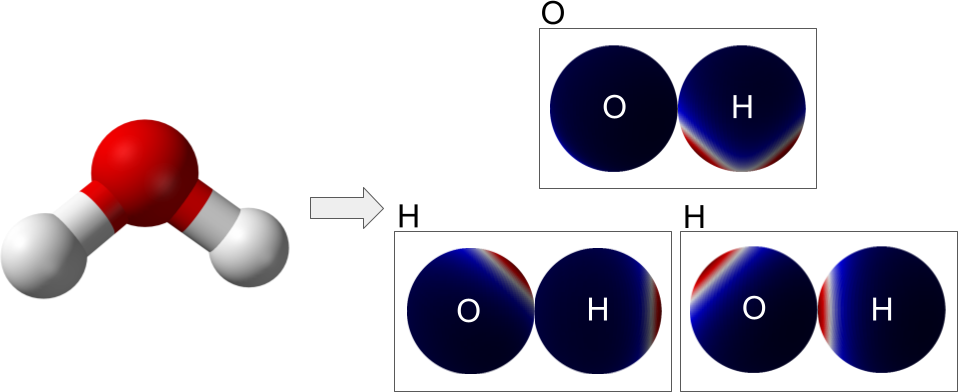}}
    \caption{ We represent a molecule with a set of $Z$ functions on
      the sphere for each atom, where $Z$ is the number of atom types
      in the dataset. Consider the $\mathrm{H_2O}$ molecule in the figure and
      let $Z=2$; the rectangles show the two channels for each
      atom. The values on the sphere come from physically-based
      interactions between pairs of atoms, smoothed with a Gaussian kernel, and aggregated over atom
      types. For example, the sphere marked with an $H$ on the top
      right sums up the Coulomb forces between the oxygen 
      the two hydrogen atoms.
      \vspace{-30pt}
    }
    \label{fig:mol2sph}
  \end{center}
\end{figure}

Thus, a molecule with $N$ atoms in a dataset containing $Z$ different
atom types is represented by $2NZ$ feature maps. The input
representation contains global information since it aggregates
interactions between all atoms, however the power law makes it biased
towards nearby atoms. \cref{fig:mol2sph} depicts the spherical
representation of an $\mathrm{H_2O}$ molecule.

This representation is computed on-device using JAX primitives
and thus is differentiable, enabling future applications such as
predicting molecule deformations or interactions.

\subsubsection{Architecture and training}
We first apply a spherical CNN separately to the input features,
at $32 \times 32$ resolution, for each atom (up to 29 on
QM9). The model contains one standard spin-spherical
convolutional block followed by 5 residual blocks as depicted in
\cref{fig:residual} (for a total of 11 convolutional layers)
with 64 to 256 channels per layer. 
Our method's computational cost roughly scales linearly with the number of atoms.

This first step results in one feature map per atom. We then apply
global average pooling which results in a set of feature vectors, one per
atom. Two different methods are used for aggregating this set
to obtain per-molecule predictions. The first method, used
for most of the QM9 targets, applies a
DeepSets~\cite{NIPS2017_f22e4747} or PointNet~\cite{qi2017pointnet}
aggregation, similarly to \citet{s.2018spherical}.  The second method
applies a self-attention
transformer~\cite{vaswani17_atten_is_all_you_need} with four layers
and four heads, and is applied only to the polarizability $\alpha$ and
electronic spatial extent $<R^2>$, which require more refined
reasoning between the atom features for accurate prediction. It is
common in the literature to use different aggregation for these and
other targets \cite{tholke2021equivariant,pmlr-v139-schutt21a}.

We train for 2000 epochs on 16 TPUv4 with batch size 16; training runs
at around 37 steps/s. 

\subsubsection{Results}
\label{sec:qm9}
\cref{tab:qm9} shows our results on the QM9 dataset. There are two
different splits used in the literature, the major difference being
that ``Split 1'' uses a training set of \num{110000} elements while ``Split
2'' uses \num{100000}. We evaluate our model on both splits and compare
against the relevant models. Our model outperforms the baselines on 8
out of 12 targets in ``Split 1'' and 9 out of 12 targets in ``Split
2''.

\subsection{Weather forecasting}
We now analyze large spherical CNNs for weather forecasting. A unique challenge here is that accurate recordings of weather data are limited to the last few decades, and thus the limited training data motivates a search for the right inductive biases for best generalization.

One potential issue to consider is that the Earth has specific
topography and orientation in space which influence the weather, and input atmospheric data is always aligned, so one could argue that global rotation equivariance 
is unnecessary or even harmful. We claim, however, that equivariance
is not harmful because we can simply include constant feature channels
such as the latitude, longitude, land-sea mask and orography at each
point. In fact, current neural weather models (NWMs) do include these
constants~\cite{rasp21_data_medium_weath_predic_with,GlobalExtremeHeatForecastingUsingNeuralWeatherModels,keisler22_forec_global_weath_with_graph_neural_networ}.

Furthermore, we speculate that rotation equivariance can be
beneficial, not in the global sense since inputs are aligned, but in
the local sense where local patterns can appear at different
orientations at different locations.

We evaluate large spherical CNNs on different settings using ERA5 reanalysis data~\cite{era5},
which combines meteorological observations with simulation models to provide
atmospheric data such as wind speed and temperatures uniformly sampled in time and space.

\subsubsection{WeatherBench}
\label{sec:wb}
The WeatherBench~\cite{rasp20_weath} benchmark
is based on ERA5 data, where the data is provided in hourly
timesteps for 40 years, for a total of around \num{350000} examples.
The dataset is accompanied by simple baseline models for predicting
geopotential height at \qty{500}{hPa} (Z500) and temperature at \qty{850}
hPa (T850) at $t+3$ days and $t+5$ days given the values at $t$.
In follow-up work, \citet{rasp21_data_medium_weath_predic_with} applied
deep residual networks to similar targets, but now taking a much larger
number of predictors including geopotential, wind speed, specific
humidity at 7 vertical levels, temperature at \qty{2}{m} (T2M), total
precipitation, and solar radiation. Each predictor is sampled at
$t$, $t-6$h and $t-12$h, and the constants land-sea mask,
orography, and latitude are included as features,
for a total of \num{117} channels.
In both settings, the inputs are sampled at $32 \timesnarrow 64$ resolution.

\textbf{Architecture and training.}
We follow \citet{rasp21_data_medium_weath_predic_with} and train a
spherical CNN with an initial block followed by 19 residual blocks
(as in \cref{fig:residual}) and no pooling, for a total of 39
spherical convolutional layers, all with 128 channels. We train one model
to directly predict Z500, T850 and T2M at 3 days ahead, and another to
predict 5 days ahead.
We train for 4 epochs on 16 TPUv4 with batch size 32; training runs
at around 8.9 steps/s. 

\textbf{Results.}
\cref{tab:wb} shows the results on the test set which comprises
years 2017 and 2018. We outperform the baseline on all metrics in the
simpler setting that takes two predictors, and show lower temperature
errors on the second setting with \num{117} predictors. The spherical
CNN even outperforms models that are pre-trained on large amounts of
simulated data on some metrics.  We notice that
models tend to overfit, and
\citet{rasp21_data_medium_weath_predic_with} employ dropout and
learning rate decay based on validation loss to mitigate the issue. We
did not use these methods, which might explain our underperforming the
geopotential target.

\begin{table}[h]
  \caption{WeatherBench results. We report the RMSE on geopotential
    height (Z500) and temperature at two verticals (T850 and T2M). The
    top block follows the protocol from \citet{rasp20_weath}, the
    middle follows \citet{rasp21_data_medium_weath_predic_with}. A
    ``cont'' superscript indicates a continuous model that takes the
    lead time as input. Spherical CNNs generally outperform
    conventional CNNs on this task, and even outperform models
    pre-trained (superscript ``pre'') on large amounts of simulated data on most temperature
    metrics.
    \vspace{-10pt}
  }
  \label{tab:wb}
  \begin{center}
  \resizebox{\columnwidth}{!}{
    \begin{tabular}{@{}lc
      S[table-format=3.0,table-auto-round]
      S[table-format=1.2,table-auto-round]
      S[table-format=1.2,table-auto-round]      
      c
      S[table-format=3.0,table-auto-round]
      S[table-format=1.2,table-auto-round]
      S[table-format=1.2,table-auto-round]      
      }
      \toprule
                                                                                  &  &               & {3 days}  &           &  &               & {5 days}  &           \\
      \cmidrule{3-5}\cmidrule{7-9}
                                                                                  &  & {Z500}        & {T850}    & {T2M}     &  & {Z500}        & {T850}    & {T2M}     \\
                                                                                  &  & {[$m^2/s^2$]} & {[K]}     & {[K]}     &  & {[$m^2/s^2$]} & {[K]}     & {[K]}     \\      
      \midrule
      \phantom{M} \emph{2 predictors}                                                              &  &               &           &           &  &               &           &           \\      
      \citet{rasp20_weath}                                                   &  & \bgl 626      & \bgl 2.87 & {-}       &  & \bgl 757      & \bgl 3.37 & {-}       \\
      Ours                                                                        &  & \bgd 531      & \bgd 2.38 & {-}       &  & \bgd 717      & \bgd 3.03 & {-}       \\      
      \midrule
      \phantom{M} \emph{117 predictors}                                                               &  &               &           &           &  &               &           &           \\      
      \citeauthor{rasp21_data_medium_weath_predic_with}\textsuperscript{cont}     &  & 331           & 1.87      & 1.60      &  & \bgd 545      & \bgd 2.57 & \bgl 2.06 \\
      \citeauthor{rasp21_data_medium_weath_predic_with}                           &  & \bgd 314      & \bgl 1.79 & \bgl 1.53 &  & \bgl 561      & \bgl 2.82 & 2.32      \\
      Ours                                                                        &  & \bgl 329      & \bgd 1.62 & \bgd 1.29 &  & 601           & \bgd 2.57 & \bgd 1.89 \\
      \midrule
      \phantom{M} \emph{Pretrained}                                                                 &  &               &           &           &  &               &           &           \\            
      \citeauthor{rasp21_data_medium_weath_predic_with}\textsuperscript{pre}      &  & 268           & 1.65      & 1.42      &  & 523           & 2.52      & 2.03      \\
      \citeauthor{rasp21_data_medium_weath_predic_with}\textsuperscript{pre,cont} &  & 284           & 1.72      & 1.48      &  & 499           & 2.41      & 1.92      \\
      \bottomrule
    \end{tabular}
  }    
\end{center}
\end{table}
  
\subsubsection{Global temperature forecasting}
\label{sec:globaltemp}
\citet{GlobalExtremeHeatForecastingUsingNeuralWeatherModels} proposed
the task of extreme heat forecasting from short to subseasonal (up to
28 days head) timescales. Current physics-based weather models cannot
forecast such long lead times, which motivates the use of machine
learning.
In contrast with \citet{rasp20_weath} and \citet{rasp21_data_medium_weath_predic_with},  \citet{GlobalExtremeHeatForecastingUsingNeuralWeatherModels} does consider
the spherical topology and employ an approximately uniform cubical sampling
on the sphere.

Data used for this task is averaged over \qty{24}{h}, and sampled
daily, resulting in around \num{15000} examples. Furthermore, data
that is not present in WeatherBench is used, such as soil moisture,
longwave radiation and vorticity. For the task we consider,
\citet{GlobalExtremeHeatForecastingUsingNeuralWeatherModels} used \num{20}
predictors while we use only the \num{5} that are present in WeatherBench, namely temperature at \qty{2}{m} (T2M), geopotential height at \qtylist{300;500;700}{hPa} and incoming radiation.
\citet{GlobalExtremeHeatForecastingUsingNeuralWeatherModels} applied a
UNet-like ~\cite{RonnebergerFB15} model on a $6 \timesnarrow 48 \timesnarrow 48$ cubemap to
forecast 28 channels corresponding to 1 to 28 days T2M.

\textbf{Architecture and training.}
We used WeatherBench data at $128 \timesnarrow 128$ resolution,
which has similar number of samples to the $6 \timesnarrow 48 \timesnarrow 48$ cubemap.
We implement a spherical UNet with 9 spherical convolutional layers with 128 channels each.
We train for 5 epochs on 16 TPUv4 with batch size 32; training runs
at around 13 steps/s.%

\textbf{Results.}
\cref{tab:heat} shows a comparison against 3 models introduced by
\citet{GlobalExtremeHeatForecastingUsingNeuralWeatherModels}
over the test set (\num{2017} to \num{2021}). HeatNet
has a loss biased towards high temperatures, ExtNet is biased towards
both hot and cold extremes, while GenNet uses a standard L2 loss like
our model. Our model nearly matches GenNet's performance, even when
using a small subset of the predictors.

\begin{table}[h]
  \caption{Temperature at \qty{2}{m} (T2M) prediction, following the
    protocol and comparing against baselines from
    \citet{GlobalExtremeHeatForecastingUsingNeuralWeatherModels}.  Our
    model nearly matches the best baseline performance, even when
    using only a small subset of predictors.}
  \label{tab:heat}
  \begin{center}
    \resizebox{\columnwidth}{!}{
      \begin{tabular}{lc
      S[table-format=2.0,table-auto-round]
      S[table-format=1.2,table-auto-round]
      S[table-format=1.2,table-auto-round]
      S[table-format=1.2,table-auto-round]      
      S[table-format=1.2,table-auto-round]
      S[table-format=1.2,table-auto-round]
      S[table-format=1.2,table-auto-round]      
      }
      \toprule
              &  &              & \multicolumn{6}{c}{T2M RMSE [K]}                                                                    \\
              &  & {Predictors} & {1 day}        & {2 days}       & {4 days}       & {7 days}       & {14 days}      & {28 days}      \\
      \midrule
      ExtNet  &  & 20           & \bgl 1.1481676 & 1.6437441      & 2.1091464      & 2.311601       & 2.400535       & \bgl 2.4156785 \\
      HeatNet &  & 20           & 1.260348       & 1.7672575      & 2.2301385      & 2.4213269      & 2.5001082      & 2.52823        \\
      GenNet  &  & 20           & \bgd 1.1297232 & \bgd 1.5992303 & \bgd 2.0277622 & \bgd 2.2207775 & \bgd 2.3105392 & \bgd 2.3392556 \\
      Ours    &  & 5            & 1.23981        & \bgl 1.63227   & \bgl 2.04294   & \bgl 2.26741   & \bgl 2.39066   & 2.45523        \\      
      \bottomrule
    \end{tabular}
    }
  \end{center}
\end{table}

\subsubsection{Iterative high resolution forecasting}
\label{sec:keisler}
\citet{keisler22_forec_global_weath_with_graph_neural_networ} proposed
an iterative graph neural network for weather forecasting. In this
setting, predictors and targets have the same cardinality such that
the model can be iterated repeatedly to forecast longer ranges, where
a single iteration produces the forecast \qty{6}{h} ahead.
Temperature, geopotential height, specific humidity and the three
components of the wind speed, are all sampled at 13 vertical levels, for a
total of 78 predictors and targets, at $180 \timesnarrow 360$ resolution. The
dataset comprises the years 1979 to 2020 with one example every
\qty{3}{h}, for a total of \num{120000} examples.

\textbf{Architecture and training.}
We use the same data as
\citet{keisler22_forec_global_weath_with_graph_neural_networ}, but at
$256 \times 256$  resolution, which has approximately the same number of samples
as the baseline.

We implement a spherical UNet with 7 convolutional layers
and a single round of subsampling, because a too large receptive
field should not be necessary for predicting \qty{6}{h} ahead
iteratively. The model is repeated up to 12 steps during training,
for a total of 84 convolutional layers. 

We train in 3 stages. The first uses 4 rollout steps (\qty{24}{h}) for 50
epochs on 32 TPUv4 with batch size 32, and is followed by 10 epochs
with 8 rollout steps (\qty{48}{h}) and 10 epochs with 12 rollout steps (\qty{72}{h}).
Training runs at around \qtylist{0.92;0.35;0.24}{steps/s} for each stage.

\textbf{Results.}
\cref{tab:keisler} and \cref{fig:qzt} show the results.
Our model tends to perform better at
geopotential but worse at temperature forecasts.
\citet{keisler22_forec_global_weath_with_graph_neural_networ} employs
a different training procedure where the resolution
changes in each stage, and the results are smoothed during evaluation.

\begin{table}[h]
  \caption{ Iterative weather forecasting, following the protocol from
    \citet{keisler22_forec_global_weath_with_graph_neural_networ}. We
    compare results for 24h, 72h and 120h lead
    times, and report the RMSE.
    Our model tends to perform better at geopotential but worse at the
    temperature forecasts.
    \vspace{-10pt}
  }
\label{tab:keisler}
  \begin{center}
    \resizebox{\columnwidth}{!}{
      \begin{tabular}{@{}
        ll
        S[table-format=2.1,table-auto-round]
        S[table-format=1.3,table-auto-round]
        l
        S[table-format=3.1,table-auto-round]
        S[table-format=1.2,table-auto-round]
        l
        S[table-format=3.1,table-auto-round]
        S[table-format=1.2,table-auto-round]
        }
        \hline
        &  & \multicolumn{2}{c}{1 day} &  & \multicolumn{2}{c}{3 days} &  & \multicolumn{2}{c}{5 days}\\
        &  & {Z500} & {T850} &  & {Z500} & {T850} &  & {Z500} & {T850}\\
        \hline
        \citet{keisler22_forec_global_weath_with_graph_neural_networ} &  & 64.89 & \bgd 0.7299 &  & 175.53 & \bgd 1.1724 &  & 344.69 & \bgd 1.7792\\
         Ours & & \bgd 58.275 & 0.82657 & & \bgd 167.19 & 1.2598 & & \bgd 340.02 & 1.9056 \\
        \hline
      \end{tabular}      
    }
  \end{center}
\end{table}

\subsection{Ablations}
\textbf{Activation, pooling, molecule representation.}
We train a small model with five spherical convolutional layers to regress
the enthalpy of atomization H on QM9; it is trained for 250 epochs (in
contrast with 2000 epochs in \cref{sec:molecules_experiments}).  We
compare the effect of replacing each of our main contributions and
show that each of them increases performance. Specifically, we compare
against the magnitude activation used in \citet{EstevesMD20},
the gated activation introduced by \citet{weiler20183d}, which we
implement by learning a spin 0 feature map that is squashed
and pointwise-multiplies each channel. We also compare against the
spherical molecular representation introduced by \citet{s.2018spherical}.
\Cref{tab:ablation_molecules} shows the results.
\begin{table}[h]
  \caption{
    Effects of activation, pooling, molecule representation.
    We employ a phase collapse activation, compared against
    the gated nonlinearity of \citet{weiler20183d} and the magnitude
    thresholding of \citet{EstevesMD20}.
    We employ spectral pooling, compared against the spatial pooling
    from \citet{EstevesMD20}.
    We introduce a novel spherical representation of molecules, compared
    against the one by \citet{s.2018spherical}.
  }
  \label{tab:ablation_molecules}
  \begin{center}
    \small{
      \begin{tabular}{@{}lllc@{}}
        \toprule
        \multirow{2}{*}{Activation} & \multirow{2}{*}{Pooling} & \quad Molecule                     & QM9/H     \\
                                    &                          & representation               & MAE (meV) \\        
        \midrule
        Ours                        & Ours                     & Ours                         & 15.25     \\
        \midrule
        Ours                        & \citeauthor{EstevesMD20}               & Ours                         & 16.13     \\
        \citeauthor{weiler20183d}   & Ours                     & Ours                         & 16.70     \\        
        \citeauthor{EstevesMD20}    & Ours                     & Ours                         & 17.01     \\
        Ours                        & Ours                     & \citeauthor{s.2018spherical} & 20.90     \\
        \bottomrule
      \end{tabular}
      }
    \end{center}
    \vspace{-15pt}
\end{table}

\textbf{Effects of scaling.}
In this experiment, we investigate how the resolutions and model
capacity affect accuracy in weather forecasting.  As in
\cref{sec:keisler}, we follow the protocol of
\citet{keisler22_forec_global_weath_with_graph_neural_networ}, but
only supervising and evaluating the forecasts \qty{6}{h} in the
future.  \Cref{tab:ablation_scaling} shows the results; the most
important factors for high performance in this task are the input and
feature maps resolutions.

\begin{table}[h]
  \caption{
    Effects of scaling. We report the RMSE for geopotential at
\qty{500}{hPa} (Z500) and temperature at \qty{850}{hPa} (T850),
predicting \qty{6}{h} ahead following
\citet{keisler22_forec_global_weath_with_graph_neural_networ}. Top row
shows our base model.  The next block reduces the input resolution.
The following row uses separable convolutions in every other layer,
which reduces the number of convolutions but keeps the feature size
constant.  The final block reduces the number of channels per layer,
which reduces both the number of operations and feature size.
  }
  \label{tab:ablation_scaling}
  \begin{center}
    \resizebox{\columnwidth}{!}{
\begin{tabular}{ccccccc}
\bottomrule
                      & \multirow{2}{*}{channels} & \multirow{2}{*}{convolutions} & \multirow{2}{*}{feature size} &  & Z500                                        & T850 \\
                      &  &  &          &  & [m\textsuperscript{2}/s\textsuperscript{2}] & [K]  \\  
\midrule
$256\timesnarrow 256$ & 100\%    & \num{3.0e5} & \num{8.4e7}          &  & 34.93                                       & 0.62 \\
\midrule
$192\timesnarrow 192$ & 100\%    & \num{3.0e5}         & \num{4.7e7}          &  & 39.68                                       & 0.74 \\
$128\timesnarrow 128$ & 100\%    & \num{3.0e5}         & \num{2.1e7}           &  & 46.39                                       & 0.87 \\
$64\timesnarrow 64$   & 100\%    & \num{3.0e5}         & \num{5.2e6}           &  & 69.60                                       & 1.08 \\
\midrule
$256\timesnarrow 256$ & 100\%    & \num{1.6e5}         & \num{8.4e7}          &  & 36.24                                       & 0.65 \\
\midrule
$256\timesnarrow 256$ & 75\%     & \num{1.7e5}         & \num{6.3e6}          &  & 36.65                                       & 0.65 \\
$256\timesnarrow 256$ & 50\%     & \num{7.4e4}          & \num{4.2e6}          &  & 41.34                                       & 0.71 \\
\bottomrule
\end{tabular}
      }
  \end{center}
\end{table}

\section{Discussion}
\textbf{Limitations.}
The major limitation of our models is still the computational cost --
our best results require training up to 4 days on 32 TPUv4,
which can be expensive. 
As a comparison, the baseline for weather \cite{keisler22_forec_global_weath_with_graph_neural_networ} trains in 5.5 days on a single GPU,
and the baseline for molecules \cite{liao22_equif} trains in 3 days on a single GPU.

From the point of view of the applications, there are concerns of how much a model trained on reanalysis is useful for forecasting the real weather, and whether models supervised by chemical properties at some level of theory like QM9 are useful to estimate the true properties.

\textbf{Conclusion.}
Spherical CNNs possess numerous qualities that makes them appealing for modeling spherical data or rotational symmetries. We
have introduced an approach to scaling these models so they
can be utilized on larger problems, and our initial results
already reach state of the art or comparable performance
on molecule and weather prediction tasks. We hope this
work and supporting implementation will allow the research community to revisit this powerful class of models
for important large scale problems.

\section*{Acknowledgments}
We thank Stephan Hoyer, Stephan Rasp, and Ignacio Lopez-Gomez for
helping with data processing and evaluation, and Fei Sha, Vivian Yang,
Anudhyan Boral, Leonardo Zepeda-Núñez, and Avram Hershko for
suggestions and discussions.

\bibliography{main}
\bibliographystyle{icml2023}

\newpage
\appendix
\twocolumn
\section{Appendix}

\subsection{Experimental details}
We use the Adam~\cite{kingma14_adam} optimizer and a cosine decay on
the learning rate with one epoch linear warmup in all experiments.

The inputs of all models are conventional spherical functions
(zero spin). The first layer maps it to features of spins zero and one,
which are mapped back to spin zero at the last spherical convolutional layer.
This last feature is complex-valued, which we convert to real by taking
the magnitude.

\subsubsection{Molecular property regression}
For the experiments in \cref{sec:molecules_experiments}, we use five
spherical residual blocks with resolutions $[32^2, 16^2, 16^2, 8^2,
8^2]$ and $[64, 128, 128, 256, 256]$ channels per layer.  We minimize
the L1 loss with a maximum learning rate of $10^{-4}$.

Our model applies the spherical CNN independently to each atom's features,
followed by global average pooling, resulting in one feature vector per atom.
These are further processed by a DeepSets or transformer, as explained in
\cref{sec:molecules_experiments}.
Finally, we map the set of atom feature vectors to the regression target
in three different ways, depending on the target.
The dipole moment $\mu$ relates to the displacement between atoms and the center of mass,
so we use a weighted average by the displacements to aggregate the atom features
(as \citet{C7SC02267K}), followed by a small MLP.
We compute the electronic spatial extent $<R^2>$ similarly, but using the distance to
the center of mass squared as the weights, following~\citet{pmlr-v139-schutt21a}.
For the other targets, which are energy-related, we use the atom types as the weights. 

Following~\citet{gasteiger20_direc_messag_passin_molec_graph}, we
estimate $\epsilon_{\text{gap}}$ as $\epsilon_{\text{HOMO}} - \epsilon_{\text{LUMO}}$, using
the predictions from models the trained for $\epsilon_{\text{HOMO}}$ and
$\epsilon_{\text{LUMO}}$, without training a model specifically for the gap.

\subsubsection{Iterative high resolution weather forecasting}
We implement a spherical UNet similar to the one in \cref{supp:globaltemp},
with feature maps of resolutions
$[256^2, 256^2, 128^2, 128^2, 128^2, 128^2, 256^2, 256^2]$ and
$[128, 128, 256, 256, 256, 256, 128, 128]$ channels per 
layer, which are followed by batch normalization and phase collapse

Similarly to~\citet{keisler22_forec_global_weath_with_graph_neural_networ},
we concatenate a few constant fields to the 78 predictors; namely,
the orography, land-sea mask, latitude (sine), longitude (sine and cosine),
hour of the year, and hour of the day. 

The maximum learning rate for the first stage is \num{2e-4},
and we reduced it by a factor of 10 at each subsequent state.

\subsubsection{WeatherBench}
For the experiments in \cref{sec:wb}, we use $64 \timesnarrow 64$
inputs and feature maps, while the baseline is at $32 \timesnarrow
64$.  Since the spherical harmonic transform algorithm we use requires
the same number of samples along both axes, we upsample the inputs
from $32 \timesnarrow 64$ to $64 \timesnarrow 64$.
We minimize the L2 loss for this and all weather experiments.

\subsubsection{Global temperature forecasting}
\label{supp:globaltemp}
For the experiments in \cref{sec:globaltemp},
we implement a spherical UNet with feature maps of resolutions
$[128^2, 64^2, 64^2, 32^2, 32^2, 32^2, 32^2, 64^2, 64^2, 128^2,
128^2]$, and 128 channels on all convolutional layers, which are
followed by batch normalization and phase collapse activation.
Features in the downsampling layers are concatenated to
the same resolutions in the upsampling layers.

\subsection{Extra experiments}
\textbf{FFT vs DFT.}
One of our perhaps surprising findings is that computing Fourier
transforms via DFT matrix multiplication is faster than using the fast
Fourier transform (FFT) algorithm. Here, we investigate whether this
remains true for larger input resolutions.  We train shallow models
with only two spin-spherical convolutional layers on the protocol of
\citet{keisler22_forec_global_weath_with_graph_neural_networ}, with
upsampled inputs to $512\times 512$ and $768\times 768$.  \Cref{tab:timing_fft}
shows the results when running on 32 TPUv4 with batch size of one per
device.  The direct DFT method performs better than FFT on TPU even at
higher resolutions, due the TPU greatly favoring computing a large
matrix multiplication instead of running multiple steps on smaller
inputs.

\begin{table}[h]
  \caption{
    Training time comparison of a shallow model using DFT and FFT for
Fourier transform computation, varying the input resolution.
      }
  \label{tab:timing_fft}
  \begin{center}
    \begin{tabular}{lllc}
      \hline
      FT method & resolution &  & steps/s\\
      \hline
      DFT & $256\times 256$ &  & 28.5\\
      FFT & $256\times 256$ &  & 18.7\\
      \hline
      DFT & $512\times 512$ &  & 12.6\\
      FFT & $512\times 512$ &  & 5.8\\
      \hline
      DFT & $768\times 768$ &  & 5.1\\
      FFT & $768\times 768$ &  & 1.8\\
      \hline
    \end{tabular}
  \end{center}
\end{table}

\textbf{TPUs vs GPUs}
While we made design decisions with TPUs in mind, the model can also
run on GPUs.  We evaluated our model for molecules
(\cref{sec:molecules_experiments}) on 8 V100 GPUs, with batch size of
1 per device, and it trains at 13.1 steps/s.  In comparison, the same
model trains at 35.6 steps/s on 16 TPUv4.

\begin{figure*}[h]
  \begin{center}
    \centerline{\includegraphics[width=\textwidth]{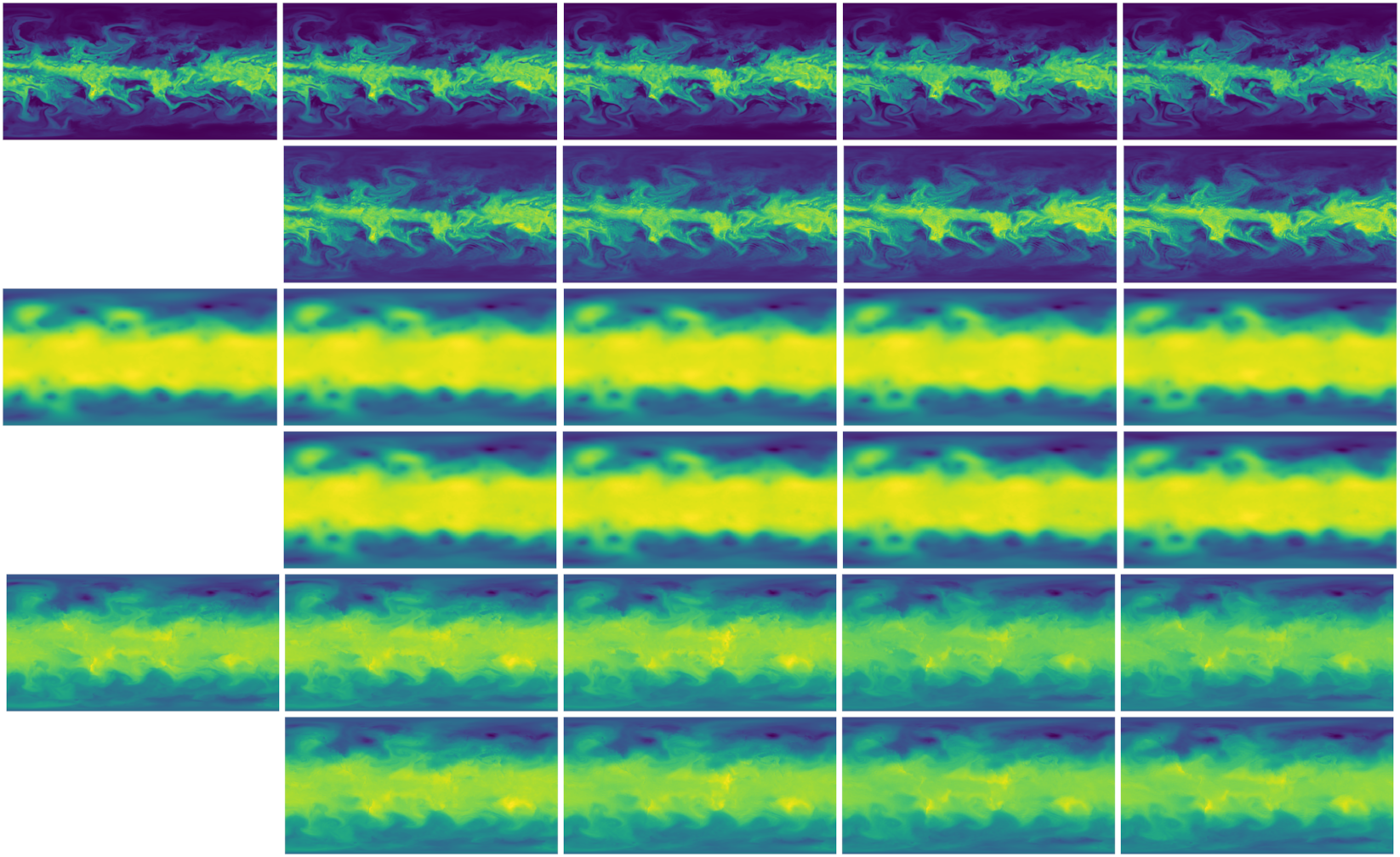}}
    \caption{One day rollout of a few predictions of our
      model.
      \emph{Top two rows:} specific humidity at \qty{850}{hPa} (Q850).
      \emph{Middle two rows:} geopotential height at \qty{500}{hPa} (Z500).
      \emph{Bottom two rows:} temperature at \qty{850}{hPa} (T500).      
      The first column shows the input values at $t=0$, and subsequent columns shows \qty{6}{h} steps.
      On each group of two rows, the top shows the ground truth
      and the bottom one shows our predictions.
      Our predictions show that large spherical CNNs are
      capable of producing high resolution outputs with high frequency
      details.}
    \label{fig:qzt}
  \end{center}
\end{figure*}

\textbf{Visualization}
\Cref{fig:qzt} shows a sequence of predictions of our model for a few variables,
compared to the ground truth.

\end{document}